\documentclass[11pt,a4paper]{article}
\usepackage[hyperref]{emnlp2020}
\usepackage{times}
\usepackage{latexsym}

\usepackage{graphicx}

\usepackage{microtype}

\aclfinalcopy % Uncomment this line for the final submission
\title{Domain specific BERT representation for Named Entity Recognition of lab protocol}

\author{Tejas Vaidhya \and Ayush Kaushal\ \\
  Indian Institute of Technology, Kharagpur \\
  \texttt{iamtejasvaidhya@gmail.com, ayushk4@gmail.com} \\}

\date{}

\begin{document}
\maketitle
\begin{abstract}
Supervised models trained to predict properties from representations, have been achieving high accuracy on a variety of tasks. For instance, the BERT family seems to work exceptionally well on the downstream task from NER tagging to the range of other linguistic tasks. But the vocabulary used in the medical field contains a lot of different tokens used only in the medical industry such as the name of different diseases, devices, organisms, medicines, etc. that makes it difficult for traditional BERT model to create contextualized embedding. In this paper, we are going to illustrate the \textbf{System for Named Entity Tagging based on Bio-Bert}.
% We are also going to provide\textbf{ Error Analysis} of our model on the given dataset at\textit{ W-NUT 2020 Shared Task-1, Named Entity Recognition}.
Experimental results show that our model gives substantial improvements over the baseline and stood the \textbf{fourth runner up} in terms of F1 score, and \textbf{first runner up} in terms of Recall with just 2.21 \textit{F1} score behind the best one.\footnote{\url{ https://github.com/tejasvaidhyadev/NER_Lab_Protocols}}

\end{abstract}
\section{Introduction}
A large amount of data is generated every year in the medical field. One of the most important generated data is the documentation of protocols. It provides individual sets of instructions that allow scientists to recreate experiments in their own laboratory. Most of them are written in Natural language which reduces its machine readability. The protocol gives a concise overview of the project which reduces its pre-processing needs but also make it less informative syntactically that eventually results in less accuracy. 

Recent progress in Named Entity Recognition was made possible by the advancements of deep learning techniques used in natural language processing (NLP). For instance, Long Short-Term Memory (LSTM) \cite{10.1162/neco.1997.9.8.1735} and Conditional Random Field (CRF) \cite{8244665}{} have greatly improved performance in biomedical named entity recognition (NER) over the last few years. Bio-BERT\citep{10.1093/bioinformatics/btz682} outperform all the other previous approaches with the help of BERT \citep{devlin2018pretraining} architecture pre-trained on Bio-medical texts \citep{10.1093/bioinformatics/bty449, 10.1093/bioinformatics/btx228, 10.1093/bioinformatics/bty869, article}.

In this paper, we are introducing our system for the NER tagging on the WLP dataset \citep{kulkarni2018wetlab}. We use a variant of Bio-Bert \citep{10.1093/bioinformatics/btz682}. The primary motivation to use the model is it's medical vocabulary and features encoded in the pre-trained model.

% \begin{table*}
% \centering
% \begin{tabular}{lll}
\begin{figure*}
\centering
   \includegraphics[width=\textwidth]{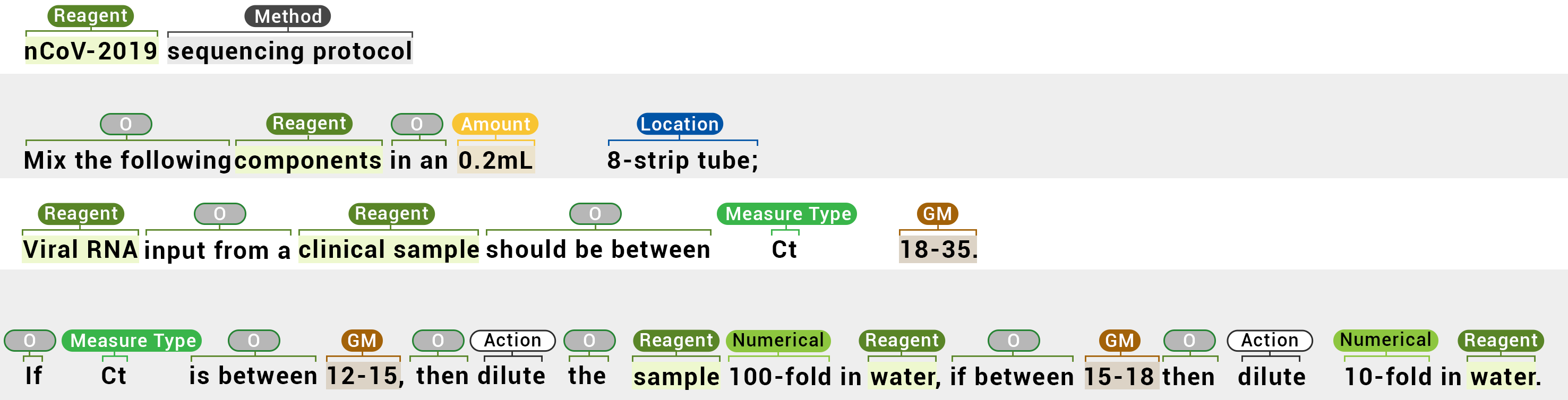}
% \end{tabular}
\caption{Visualisation of annotated dataset \footnotemark}
\label{data-formate}
\end{figure*}

% \end{table*}

\section{Task Description and Data Set}

Formally, the WNUT 2020 Shared Task-1 Named Entity Recognition, organized within, the 6th Workshop on Noisy User-generated Text (WNUT), 2020 \cite{tabassum2020wlp} is a NER prediction task. It can be expressed as `tokens-level' classification task mathematically as:

Let the sentence \(S\) be defined as:

$S = \{ s_1, s_2, ..., s_n \}$

$n$ is the number words in the sentences can be classified into the following label set 

$y = \{l_1,l_2,l_3, ..., l_m\}$

where $m$ is labels.

Given named entity of type XXX. Whenever two entities of
type XXX are immediately next to each other, the
first word of the second entity will be tagged B-XXX
in order to show that it starts another entity and the entities inside B-XXX will be represented as I-XXX. For example, the sentence  \{nCoV-2019, sequencing, protocol\} have the following labels \{B-Reagent, B-Method, I-Method\}.

\paragraph{DataSet:}  All of the protocols \cite{kulkarni2018wetlab} were collected from protocols.io using their public APIs by organising team. For the shared Task-1 W-NUT 2020: Named Entity Extraction, the annotation of 615 protocols are re-annotated using BART styled annotated protocols by 3 annotators with 0.75 inter-annotator agreement, measured by span-level Cohen's Kappa. The re-annotators incorporate missing entity-relations and also corrected the inconsistencies.

The task aims to create the system for Protocol-Named Entities Recognition (NER). The main difference that makes it difficult for traditional NER taggers is the vast vocabulary in medical filed and use of limited syntactic information. For instances "QIAprep Spin Miniprep" is device used in medical industry, but not present in our regular vocabulary that also makes it difficult for traditional NER tagger to learn.

Figure~\ref{data-formate} provides the visualization of annotated datasets provided by WNUT 2020 Shared Task-1 .

\footnotetext{Image Source: Github repository of WNUT 2020 Shared Task-1 NER.}

\section{Approach}

In the section \ref{archi} we are going to define our proposed architecture and in section  \ref{bio}  briefly review the Bio-BERT \cite{10.1093/bioinformatics/btz682} used for final submission  and also different Domain specific BERT based model used for experiment as shown in the Table \ref{result}.

\paragraph{Baseline} The organiser provided a simple Linear CRF model\footnote{\url{https://github.com/jeniyat/WNUT_2020_NER/tree/master/code/baseline_CRF}}. It utilized simple gazetteers and hand-crafted feature to predict the entities from the test data. We replaced it with our proposed  BERT based Architecture as describe in the section below.
\subsection{Architecture} \label{archi} As described in figure \ref{fig:model}, we  first sub-word tokenize each token of sentences, using BERT's wordpiece tokenizer of Huggingface library and pass it through different domain specific BERT models or BERT Transformer stacks (scibert, biobert, bert-based, bert-large etc) to extract contextualised representation \citep{Beltagy2019SciBERT,10.1093/bioinformatics/btz682,devlin2018pretraining}. We then select the representation of first sub-word token for each word and use simple Linear or Dense layer with the softmax activation function as classifier to get probability on the labels from the contextualised representation. 
\subsection{Bio-BERT} \label{bio}

%\begin{figure}
%\centering
% \centering
  %\includegraphics[scale=0.2]{prtraingbio.png}
    %\caption{overview of pretraing of BioBERT \cite{10.1093/bioinformatics/btz682}}
    %\label{fig:archi_bio}
%\end{figure}

\begin{table*}
\centering
\begin{tabular}{lllll}
\hline
\textbf{Models} & \textbf{F1-score} & \textbf{Recall} & \textbf{Precision}\\
\hline
\verb|biobert_v1.1_pubmed| & 79.10 &  79.72 & 78.61 \\ 
\verb|biobert_v1.0_pubmed_pmc|  & 79.02 &  79.51 & 79.02 \\
\verb|scibert_uncased|  & 77.66 &  79.60 & 76.00 \\
\verb|bert-large-cased|  & 77.79  &  78.74 & 77.10 \\ 
\verb|bert-large-uncased|  & 75.50 &  77.39 & 73.79 \\ 
\verb|bert-base-cased|  &  78.05 &  79.29 & 76.87  \\
\verb|Baseline |  & 74.39  &  73.32 & 75.49 \\\hline
\end{tabular}

\caption{\label{result} 
{Shows the results of test set provided by shared task organisers during experimental and details of the experimental setting is describe in section \ref{exp}}
}

\end{table*}
Bio-BERT \cite{10.1093/bioinformatics/btz682} is a contextualized language representation model, based on BERT, a pre-trained model that is trained on different combinations of general \& biomedical domain corpora.
According to \citet{10.1093/bioinformatics/btz682}, just like its parental model BERT, it is also capable of capturing contextualized bidirectional representations. Thus it has outperformed existing architectures in most of the Named Entity Recognition tasks within the biomedical domain by using a limited amount of dataset. We hypothesize that such domain-specific bidirectional representations are also critical for our task. Bio-BERT are pre-trained on the following different datasets \textit{\{Wiki + Books, Wiki + Books + PubMed, Wiki + Books + PMC, Wiki + Books + PubMed + PMC\}}.

\begin{figure}
\centering
% \centering
  \includegraphics[scale=0.30]{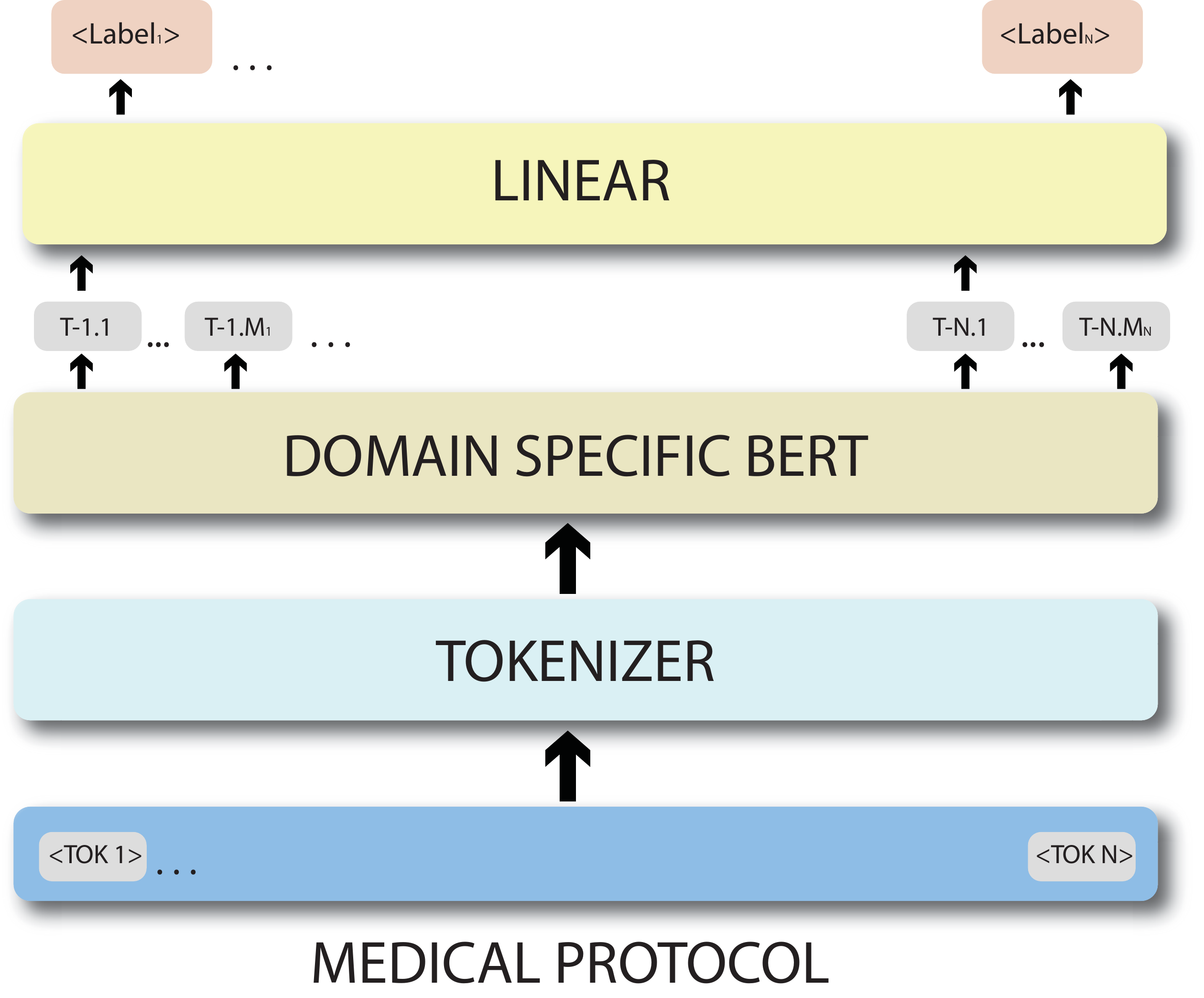}
    \caption{The TOK X represents the token of sentence where X \(\in\) N and N is the length of sentence. T-I.J represents the I\(^{th}\) tokens and J\(^{th}\) subtokens and we used WordPiece tokenizer for all of our models.}
    \label{fig:model}
\end{figure}

we again hypothesize to achieved best performance in \textit{PubMed}(comprises more than 30 million citations for biomedical literature from MEDLINE, life science journals, and online books.) trained dataset because of its linguistic similarity with protocols. For instance both of them contains medical procedure to reproduce medical experiment.

\paragraph{Other Models used in Experiment}We also used other domain specific BERT for experimentation using the same architecture as discussed in section \ref{archi} with replacement of BERT Models in place of BioBERT (as shown in figure \ref{fig:model})
\begin{itemize}
\item Sci-BERT \cite{Beltagy2019SciBERT} : A pre-trained contextualized embedding model based on BERT to address the lack of high-quality, large-scale labeled scientific data.
\item BERT$_{\{large/base, cased/uncased\}}$ :BERT \cite{devlin2018pretraining} It is designed to train deep bidirectional representations by jointly conditioning on both left and right context in all layers. Language models have demonstrated that rich, unsupervised pre-training is an integral part of many language understanding systems. Hence, we try fine-tuning BERT to obtain better results on this task.
\end{itemize}

%\cite{bring ur approach and architecture to one place, then descrive baseline models you experimented with}

\begin{table*}
\centering
\begin{tabular}{lllll}
\hline
\textbf{Models} & \textbf{F1-score} & \textbf{Recall} & \textbf{Precision}\\
\hline
\verb|biobert_v1.1_pubmed (partial match)| & 79.54 &  77.43 & 81.76 \\ 
\verb|biobert_v1.0_pubmed_pmc (complete match)|  & 74.91 &  72.93 & 77  \\\hline
\end{tabular}

\caption{\label{result-2} 
{Results on the held-out test set provided by shared task organisers on final submission}
}

\end{table*}

\section{Comparison and Discussion} \label{exp}

We experimented with different BERT models as shown in Table \ref{result}. We avoid any preprocessing other than the Bert specific tokenization, as it may result in loss of crucial semantic information in text. We also assumed protocols are relatively less noisy compare to other crowd source data with  compact sentences.
\subsection{Experimental setting}
We keep maximum length of input sentence to 512 to consider Long sentences in protocols. For all the base models (12-layer, 768-hidden, 12-heads, 110M parameters) including our Bio-BERT model, we train all for 8 epoch with batch size 16. Large models(24-layer, 1024-hidden, 16-heads, 340M parameters.) are trained for 4 epochs with batch size 16. We early stop the models using the valid set. The dropout probability was set to 0.1 for all layers. Optimization is done using Adam\cite{article1} with a learning rate of 1e-5. The remaining hyperparameters were kept same as \citet{devlin2018pretraining}. We used the PyTorch \cite{NEURIPS2019_9015} implementation of BERT from Huggingfaces tranformers \cite{Wolf2019HuggingFacesTS} library.

For selecting best models in experimental phase (i.e. before release of test set) we use split of 60/20/20 for train, dev and test respectively. Our final submission, used a 70/30 split for train/valid set of initial data and Bio-BERT\cite{10.1093/bioinformatics/btz682} model and split sentence with more than 512 tokens to two more sentence to get desired  model's  input sentence length. To evaluate the performance of the system, an evaluation script along with the dataset was provided by the organizers\footnote{\url{https://github.com/jeniyat/WNUT_2020_NER/tree/master/code/eval}}.

%\tyss{can merge this section with next and name it as Performance Comparison and Disccusion}

\subsection{Results and Inferences}

Our Bio-Bert\cite{10.1093/bioinformatics/btz682} based model performed best of all the models because of domain specific knowledge. Our model performed extremely well on final test set as shown in Table \ref{result-2} and stood \textit{\(4th\) runner} up in term of F1 score and\textit{ 1\(st\) runner up} in term of recall out of 13 teams participated in the competition.

\paragraph{Inferences:} Use of uncased large BERT results in significant loss of about 2.29 F1 score in comparison of cased-large BERT \cite{devlin2018pretraining}, clearly shows importance of syntactic nature of protocols. We observed better performance, after increase in both training and validation set for our final submissions indicating inability of model to fully capture the representation due to fine tuning on limited data.
\section{Error Analysis}
\begin{enumerate}
\item BERT tokenizer is not efficient on Bio-Medical text as illustrated in Table \ref{tokenizer}
. Its vocabulary does not consists of Bio-medical words and it is not trained on domain specific setting. Hence makes it difficult for BERT model to learn the encoding based on poorly sub tokenized word. The possible solution will be training tokenizers on both bio-medical and general text sets.
\item Treatment of task as token level classification problem results in incorrect detection of intermediate Entity as illustrated in Table \ref{err}. Hence evidenced in decrease of F1 score in complete match compare to partial match in our submission  provided by organisers.
\begin{table}
\centering
\begin{tabular}{lrr}
\hline \textbf{Tokens} & \textbf{correct labels} & \textbf{predicted labels} \\ \hline

standard & B-Reagent & B-Reagent \\
T4 & I-Reagent & B-Device \\
DNA & I-Reagent & I-Reagent \\
Ligase & I-Reagent & I-Reagent \\

\hline
\end{tabular}
\caption{\label{err} Error arises due to consideration of token level classification  }
\end{table}

\begin{table}
\centering
\begin{tabular}{ll}
\hline \textbf{Bio-Med Terms} & \textbf{subword-tokenized} \\ 

 acetyltransferase & [`ace',`ty',`lt',`ran',`s',\\
 & `fer',`ase'] \\
 Hematoxylin & [`He',`mat',`ox',`yl',`in'] \\
 sulfanilamide & [`su',`lf',`ani',`lam',`ide'] \\
 ddH2O & [`d',`d',`H',`2',`O'] \\
 lBiotin-16-UTP & [`l',`B',`iot',`in',`-',`16',\\
 & `-',`U',`TP'] \\

\hline
\end{tabular}
\caption{\label{tokenizer} Illustration of inefficient sub-tokenization of Bio-Med words  }
\end{table}

\item Use of Nomenclature, Scientific formula, abbreviations makes it difficult for Pre-trained language models to generalised with limited fine tuning data. Though with the help of contextual details it is observed that the BERT was able to correctly predict scientific formula at some places. For instance, in the sentence \textit\{Add, 5gm, SDS\},  \textit{SDS} was correctly labelled as \textit{Reagent} by our model.

\end{enumerate}

\section{Conclusion and Future Work}
In this paper, we present our system for the Named Entity recognition for Bio-medical protocols a for Shared Task at W-NUT Task-1 2020. We built upon the recent success of Pre-trained language models and
apply them for protocols. Our System achieves close to state-of-art performance on this task.

As future work, we will try to experiment with XLNet\cite{xlnet} and different ensembling between the models and would like to extend the work of  \citet{clark_et_al_2019} by performing layer by layer Analysis of  BERT.

\section*{Acknowledgments}
We would like to thank the Computer Science
and Engineering Department of Indian Institute of Technology, Kharagpur for providing us
the computational resources required for performing various experiments.We are very grateful for the invaluable suggestions given by T.Y.S.S. santosh\footnote{\url{https://www.linkedin.com/in/santosh-t-y-s-s-39227b116/}} and Aarushi Gupta. We also thank the organizers of the Shared Task-1 at WNUT, EMNLP-2020.
\bibliography{anthology,emnlp2020}

\begin{thebibliography}{16}
\expandafter\ifx\csname natexlab\endcsname\relax\def\natexlab#1{#1}\fi

\bibitem[{Beltagy et~al.(2019)Beltagy, Lo, and Cohan}]{Beltagy2019SciBERT}
Iz~Beltagy, Kyle Lo, and Arman Cohan. 2019.
\newblock \href {http://arxiv.org/abs/arXiv:1903.10676} {Scibert: Pretrained
  language model for scientific text}.
\newblock In \emph{EMNLP}.

\bibitem[{Clark et~al.(2019)Clark, Khandelwal, Levy, and
  Manning}]{clark_et_al_2019}
Kevin Clark, Urvashi Khandelwal, Omer Levy, and Christopher~D. Manning. 2019.
\newblock What does bert look at? an analysis of bert's attention.
\newblock In \emph{BlackBoxNLP@ACL}.

\bibitem[{Devlin et~al.(2018)Devlin, Chang, Lee, and
  Toutanova}]{devlin2018pretraining}
Jacob Devlin, Ming-Wei Chang, Kenton Lee, and Kristina Toutanova. 2018.
\newblock \href {http://arxiv.org/abs/1810.04805} {Bert: Pre-training of deep
  bidirectional transformers for language understanding}.
\newblock Cite arxiv:1810.04805Comment: 13 pages.

\bibitem[{Giorgi and Bader(2018)}]{10.1093/bioinformatics/bty449}
John~M Giorgi and Gary~D Bader. 2018.
\newblock \href {https://doi.org/10.1093/bioinformatics/bty449} {{Transfer
  learning for biomedical named entity recognition with neural networks}}.
\newblock \emph{Bioinformatics}, 34(23):4087--4094.

\bibitem[{Habibi et~al.(2017)Habibi, Weber, Neves, Wiegandt, and
  Leser}]{10.1093/bioinformatics/btx228}
Maryam Habibi, Leon Weber, Mariana Neves, David~Luis Wiegandt, and Ulf Leser.
  2017.
\newblock \href {https://doi.org/10.1093/bioinformatics/btx228} {{Deep learning
  with word embeddings improves biomedical named entity recognition}}.
\newblock \emph{Bioinformatics}, 33(14):i37--i48.

\bibitem[{Hochreiter and Schmidhuber(1997)}]{10.1162/neco.1997.9.8.1735}
Sepp Hochreiter and J\"{u}rgen Schmidhuber. 1997.
\newblock \href {https://doi.org/10.1162/neco.1997.9.8.1735} {Long short-term
  memory}.
\newblock \emph{Neural Comput.}, 9(8):1735–1780.

\bibitem[{Kingma and Ba(2014)}]{article1}
Diederik Kingma and Jimmy Ba. 2014.
\newblock Adam: A method for stochastic optimization.
\newblock \emph{International Conference on Learning Representations}.

\bibitem[{Kulkarni et~al.(2018)Kulkarni, Xu, Ritter, and
  Machiraju}]{kulkarni2018wetlab}
Chaitanya Kulkarni, Wei Xu, Alan Ritter, and Raghu Machiraju. 2018.
\newblock An annotated corpus for machine reading of instructions in wet lab
  protocols.
\newblock In \emph{Proceedings of the 2018 Conference of the North American
  Chapter of the Association for Computational Linguistics: Human Language
  Technologies (NAACL)}.

\bibitem[{Lee et~al.(2019)Lee, Yoon, Kim, Kim, Kim, So, and
  Kang}]{10.1093/bioinformatics/btz682}
Jinhyuk Lee, Wonjin Yoon, Sungdong Kim, Donghyeon Kim, Sunkyu Kim, Chan~Ho So,
  and Jaewoo Kang. 2019.
\newblock \href {https://doi.org/10.1093/bioinformatics/btz682} {{BioBERT: a
  pre-trained biomedical language representation model for biomedical text
  mining}}.
\newblock \emph{Bioinformatics}, 36(4):1234--1240.

\bibitem[{{Namikoshi} et~al.(2017){Namikoshi}, {Ohta}, {Takasu}, and
  {Adachi}}]{8244665}
D.~{Namikoshi}, M.~{Ohta}, A.~{Takasu}, and J.~{Adachi}. 2017.
\newblock Crf-based bibliography extraction from reference strings using a
  small amount of training data.
\newblock In \emph{2017 Twelfth International Conference on Digital Information
  Management (ICDIM)}, pages 59--64.

\bibitem[{Paszke et~al.(2019)Paszke, Gross, Massa, Lerer, Bradbury, Chanan,
  Killeen, Lin, Gimelshein, Antiga, Desmaison, Kopf, Yang, DeVito, Raison,
  Tejani, Chilamkurthy, Steiner, Fang, Bai, and Chintala}]{NEURIPS2019_9015}
Adam Paszke, Sam Gross, Francisco Massa, Adam Lerer, James Bradbury, Gregory
  Chanan, Trevor Killeen, Zeming Lin, Natalia Gimelshein, Luca Antiga, Alban
  Desmaison, Andreas Kopf, Edward Yang, Zachary DeVito, Martin Raison, Alykhan
  Tejani, Sasank Chilamkurthy, Benoit Steiner, Lu~Fang, Junjie Bai, and Soumith
  Chintala. 2019.
\newblock \href
  {http://papers.neurips.cc/paper/9015-pytorch-an-imperative-style-high-performance-deep-learning-library.pdf}
  {Pytorch: An imperative style, high-performance deep learning library}.
\newblock In \emph{Advances in Neural Information Processing Systems 32}, pages
  8024--8035. Curran Associates, Inc.

\bibitem[{Tabassum et~al.(2020)Tabassum, Xu, and Ritter}]{tabassum2020wlp}
Jeniya Tabassum, Wei Xu, and Alan Ritter. 2020.
\newblock {WNUT-2020 Task 1: Extracting Entities and Relations from Wet Lab
  Protocols}.
\newblock In \emph{Proceedings of EMNLP 2020 Workshop on Noisy User-generated
  Text (WNUT)}.

\bibitem[{Wang et~al.(2018)Wang, Zhang, Ren, Zhang, Zitnik, Shang, Langlotz,
  and Han}]{10.1093/bioinformatics/bty869}
Xuan Wang, Yu~Zhang, Xiang Ren, Yuhao Zhang, Marinka Zitnik, Jingbo Shang,
  Curtis Langlotz, and Jiawei Han. 2018.
\newblock \href {https://doi.org/10.1093/bioinformatics/bty869} {{Cross-type
  biomedical named entity recognition with deep multi-task learning}}.
\newblock \emph{Bioinformatics}, 35(10):1745--1752.

\bibitem[{Wolf et~al.(2019)Wolf, Debut, Sanh, Chaumond, Delangue, Moi, Cistac,
  Rault, Louf, Funtowicz, Davison, Shleifer, von Platen, Ma, Jernite, Plu, Xu,
  Scao, Gugger, Drame, Lhoest, and Rush}]{Wolf2019HuggingFacesTS}
Thomas Wolf, Lysandre Debut, Victor Sanh, Julien Chaumond, Clement Delangue,
  Anthony Moi, Pierric Cistac, Tim Rault, Rémi Louf, Morgan Funtowicz, Joe
  Davison, Sam Shleifer, Patrick von Platen, Clara Ma, Yacine Jernite, Julien
  Plu, Canwen Xu, Teven~Le Scao, Sylvain Gugger, Mariama Drame, Quentin Lhoest,
  and Alexander~M. Rush. 2019.
\newblock Huggingface's transformers: State-of-the-art natural language
  processing.
\newblock \emph{ArXiv}, abs/1910.03771.

\bibitem[{Yang et~al.(2019)Yang, Dai, Yang, Carbonell, Salakhutdinov, and
  Le}]{xlnet}
Zhilin Yang, Zihang Dai, Yiming Yang, Jaime Carbonell, Ruslan Salakhutdinov,
  and Quoc~V. Le. 2019.
\newblock \href {http://arxiv.org/abs/arXiv:1906.08237} {Xlnet: Generalized
  autoregressive pretraining for language understanding}.

\bibitem[{Yoon et~al.(2019)Yoon, So, Lee, and Kang}]{article}
Wonjin Yoon, Chan So, Jinhyuk Lee, and Jaewoo Kang. 2019.
\newblock \href {https://doi.org/10.1186/s12859-019-2813-6} {Collabonet:
  collaboration of deep neural networks for biomedical named entity
  recognition}.
\newblock \emph{BMC Bioinformatics}, 20:249.

\end{thebibliography}
\bibliographystyle{acl_natbib}
\end{document}